# Fast Cosine Transform to increase speed-up and efficiency of Karhunen-Loève Transform for lossy image compression

Mario Mastriani, and Juliana Gambini

*Abstract*—In this work, we present a comparison between two techniques of image compression. In the first case, the image is divided in blocks which are collected according to zig-zag scan. In the second one, we apply the Fast Cosine Transform to the image, and then the transformed image is divided in blocks which are collected according to zig-zag scan too. Later, in both cases, the Karhunen-Loève transform is applied to mentioned blocks. On the other hand, we present three new metrics based on eigenvalues for a better comparative evaluation of the techniques. Simulations show that the combined version is the best, with minor Mean Absolute Error (MAE) and Mean Squared Error (MSE), higher Peak Signal to Noise Ratio (PSNR) and better image quality. Finally, new technique was far superior to JPEG and JPEG2000.

## I. Introduction

MODERN image compression techniques often involve Discrete Cosine Transform (DCT) [1-26] with different Fast Cosine Transform (FCT) versions for a fast implementation [6,10,11,12,14,16,17] and Karhunen-Loève Transform (KLT) [27-29]. While DCT is applied to image compression [1,2,4,6, 8-12,16,17,19-21], KLT is applied in image decorrelation [30-34], that is to say, KLT is used inside compression techniques of several images with a high degree of mutual correlation, for example, frames of medical images [35], video [36, 37], and multi [30, 32-34] and hyperspectral imagery [38-40].

Many efforts have been made in the recent years in order to compress efficiently such data sets. The challenge is to have a data representation which takes into account at the same time both the advantages and disadvantages of KLT [29], for a most efficient compression based on an optimal decorrelation.

Several authors have tried to combine the DCT with the KLT but with questionable success [1], with particular interest to multispectral imagery [30, 32, 34].

In all cases, the KLT is used to decorrelate in the spectral domain. All images are first decomposed into blocks, and each block uses its own KLT instead of one single matrix for the whole image. In this paper, we use the KLT for a decorrelation between sub-blocks resulting of the applications of a DCT with zig-zag scan, that is to say, in the spectral domain.

We introduce in this paper an appropriate sequence, decorrelating first the data in the spatial domain using the DCT and afterwards in spectral domain, using the KLT, allows us a more efficient (and robust, in presence of noise) compression scheme.

The resulting compression scheme is a lossy image compression. This type of compression system does not retain the exact image pixel to pixel. Instead it takes advantage of limitations in the human eye to approximate the image so that it is visually the same as the original. These methods can achieve vastly superior compression rates than lossless methods, but they must be used sensibly [41].

Lossy compression techniques generally only work well with real-life photography; they often give disastrous results with other types of images such as line art, or text. Putting an image through several compression-decompression cycles will cause the image to deteriorate beyond acceptable standards. So a lossy compression should only be used after all processing has been done, it should not be used as an intermediate storage format. Further note that while the reconstructed image looks the same as the original, this is according to the human eye. If a computer has to process the image in a recognition system, it may be completely thrown off by the changes [41].

On the other hands, consider the generic transform coder in Fig.1 consisting of a 2-D transform, quantizer, and entropy coder. We see here that loss occurs during quantization and after the transform. Therefore, in order to conduct our analysis, we must repeat the transform to return to the stage where loss occurs and examine the effect of quantization on transform coefficients [42].

In this work, additional losses are incorporated, because, after of KLT applications a pruning of decorrelated sub-blocks is applied before the quantization, with a statistical criterion [28].

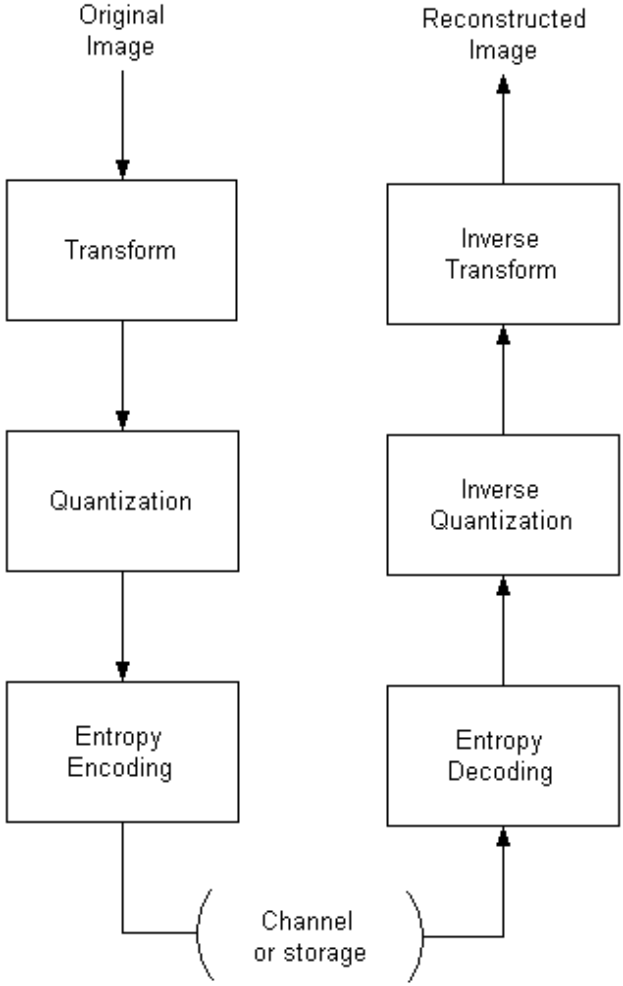

Fig. 1 Generic transform coding for digital images

The Bidimensional Discrete Cosine Transform and its fast implementation are outlined in Section II. Zig-zag scan is outlined in Section III. KLT is outlined in Section IV. Combinations are outline in Section V. In Section VI, we discuss briefly the more appropriate metrics for compression. In Section VII, the experimental results using the proposed algorithm are presented. Finally, Section VIII provides a conclusion of the paper.

## II. BIDIMENSIONAL DISCRETE COSINE TRANSFORM

The Bidimensional Discrete Cosine Transform (DCT-2D) [1,2,4,6,8-12,16,17,19-21], demostrated its superiority in front of Discrete Wavelet Transform to work in combination with the KLT in decorrelation and compression processes [43]. However, for all practical cases, it is necessary a fast implementation of the same [6,10,11,12,14,16,17].

Since the 2-D DCT (typically 4 x 4, 8 x 8 and 16 x 16) is the standard decorrelation transform in the international image/video coding standards [44] it is not suprising that research efforts have been concentrated to develop algorithms for the efficient computation of 2-D DCT only. The orthonormal 2-D DCT for an $N \times N$ input data matrix $\{x_{nm}\}$, $m, n = 0, 1, \ldots, N$-1 is defined by the following relation [44]:

$$y_{kl} = \tfrac{2}{N} \varepsilon_k \varepsilon_l \sum_{m=0}^{N-1} \sum_{n=0}^{N-1} x_{mn} \cos\left[\tfrac{\pi(2m+1)k}{2N}\right] \cos\left[\tfrac{\pi(2n+1)l}{2N}\right], \quad (1)$$

$$k, l = 0, 1, \ldots, N-1$$

and the inverse 2-D DCT, as

$$y_{kl} = \tfrac{2}{N} \sum_{k=0}^{N-1} \sum_{l=0}^{N-1} \varepsilon_k \varepsilon_l \, y_{kl} \cos\left[\tfrac{\pi(2m+1)k}{2N}\right] \cos\left[\tfrac{\pi(2n+1)l}{2N}\right], \quad (2)$$

$$m, n = 0, 1, \ldots, N-1$$

where

$$\varepsilon_p = \begin{cases} \tfrac{1}{\sqrt{2}} & p = 0, \\ 1 & otherwise. \end{cases} \quad (3)$$

Where, the signal flow graph for the forward 4x4 DCT computation, can be see on Fig.2.

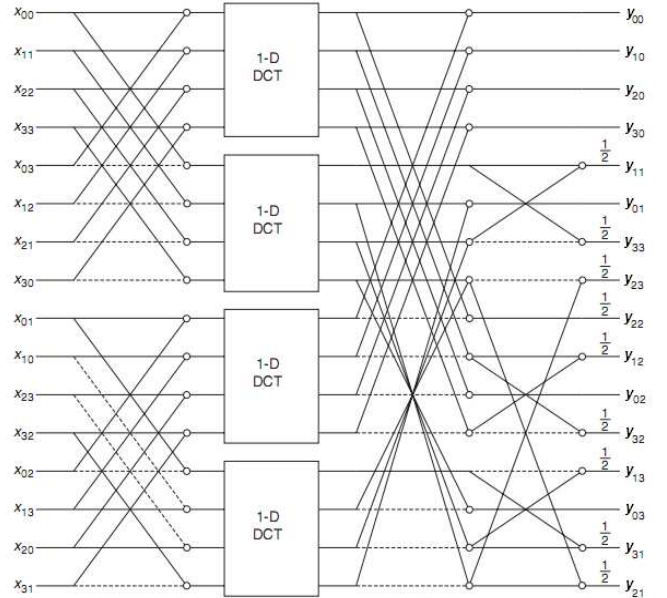

Fig. 2: The signal flow graph for the forward 4x4 DCT computation.

The fast algorithms for the direct 8×8 DCT computation [44] are derived using an algebraic and computational theoretical approach. First, a matrix factorization of DCT transform matrix (1) is converted (with additions and permutations) to a direct sum of matrices corresponding to certain polynomial products modulo irreducible polynomials. Then, these constructions using theorems regarding the structure of Kronecker products of Matrices are exploited to derive efficient 8×8 DCT algorithms. Although a practical fast Algorithm for the 8×8 DCT computation requires 94 multiplications and 454 additions, Its computational structure is rather complicated.

## III. ZIG-ZAG SCAN

Fig.3 shows the zig-zag spatial scanning method [2], which is fundamental for JPEG compression algorithm [2].

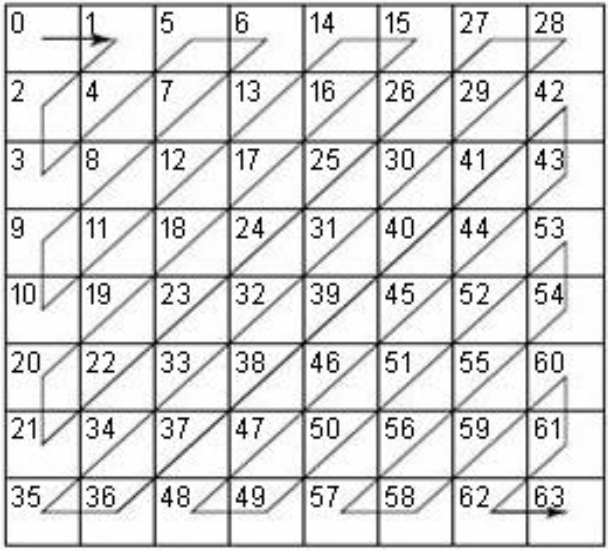

Fig. 3: Zig-zag space scanning method order.

In Fig.3 each numbering cell represent a sub-block (inside spectral domain) which may be spatially ordered (in upward order) in a three dimensional matrix before KLT, see Fig.4.

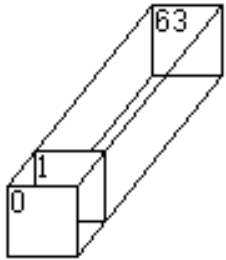

Fig. 4: Building of 3D-matrix with sub-blocks in upward order

As can be seen from Fig.3, pixels, which have to be treated or not with a DCT, are concentrated in blocks. Block clusters of 2×2, 4×4, 8×8 … pixels, can be easily extracted, since pixels in these blocks are transmitted one after another (zig-zag ordering, the same ordering employed in JPEG image compression format [2]). This feature can be handy for spatial image processing, such as resolution reduction. In order to reduce image resolution by a factor of two, the mean of four pixels (a 2×2 block) has to be calculated. With this ordering (zig-zag), it can be done in a simple, straightforward way, without requiring multiple storage elements. This calculation can be expanded to blocks of sizes 4×4, 8×8 etc.

## IV. KARHUNEN-LOEVE TRANSFORM (KLT)

The KLT begin with the covariance matrix of the vectors **x** generated between values of pixel with similar allocation in all arranged sub-blocks of 3D-matrix, as show in Fig.5.

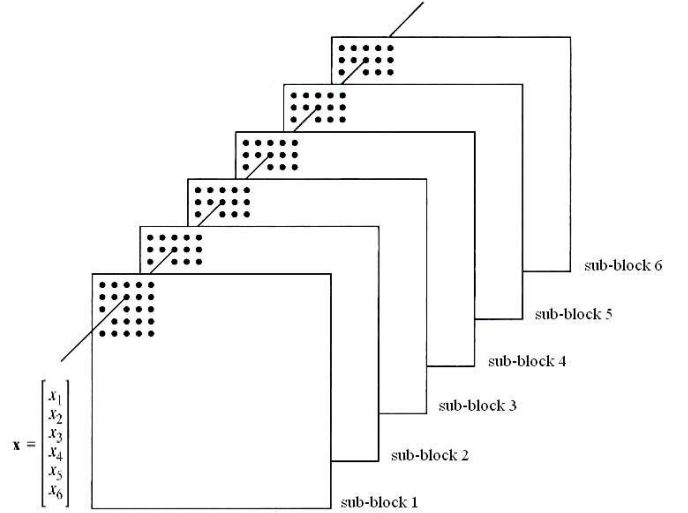

Fig. 5: Formation of a vector from corresponding pixels in six sub-blocks

The covariance matrix results,

$$\mathbf{C_x} = E\{(\mathbf{x}\text{-}\mathbf{m_x})(\mathbf{x}\text{-}\mathbf{m_x})^T\} \quad (4)$$

with:

- $\mathbf{x} = (x_1, x_2, \ldots, x_n)^T$, where **x** is one of the correlated original vector set, "$T$" indicates transpose and $n$ is the number of sub-blocks.
- $\mathbf{m_x} = E\{\mathbf{x}\}$ is the mean vector, and where $E\{\bullet\}$ is the expected value of the argument, and the subscript denotes that **m** is associated with the population of **x** vectors.

In the appropriate mathematical form:

$$m_x = \frac{1}{rsb*csb} \sum_{k=1}^{rsb*csb} x_k \quad (5)$$

where:
rsb is the sub-block row number
csb is the sub-block column number

On the other hands,

$$C_x = \frac{1}{rsb*csb} \sum_{k=1}^{rsb*csb} (x_k - m_x)(x_k - m_x)^T \quad (6)$$

Therefore, KLT will be,

$$\mathbf{y} = \mathbf{V}^T (\mathbf{x}\text{-}\mathbf{m_x}) \quad (7)$$

with:

$\mathbf{y} = (y_1, y_2, \ldots, y_n)^T$, where **y** is one of the decorrelated transformed vector set

**V** is a matrix whose columns are the eigenvectors of **C**$_x$.

When applying the calculus of eigenvectors, two matrices arise, **V** y **C**$_y$, being **C**$_y$ a diagonal matrix, where the elements on its main diagonal are de eigenvalues of **C**$_x$.

If we wish to calculate the covariance matrix of vectors **y**, results

$$\mathbf{C_y} = E\{(\mathbf{y\text{-}m_y})(\mathbf{y\text{-}m_y})^T\} = E\{\mathbf{yy}^T\} \tag{8}$$

Because, **m**$_y$ is a null vector. Besides, **C**$_y$ is a diagonal matrix. Depending on the correlation degree between the original sub-blocks, KLT will be more or less efficient decorrelating them. Such efficiency depends on how the elements of the main diagonal of the covariance matrix **C**$_y$ fall in value, from right to left. The faster they fall in value, the KLT will be more efficient decorrelating them. As an example, based on Fig.6, which represents to Lena of 512-by-512 pixels, and if we work with sub-blocks of 64-by-64 pixels, as we must see in Fig.7, we obtain the eigenvalues of Fig.8. However, if by a determined method we are starting from a set of sub-blocks as those shown in Fig.9, then we will obtain the eigenvalues of Fig.10. The second case is highly more efficient than the first one.

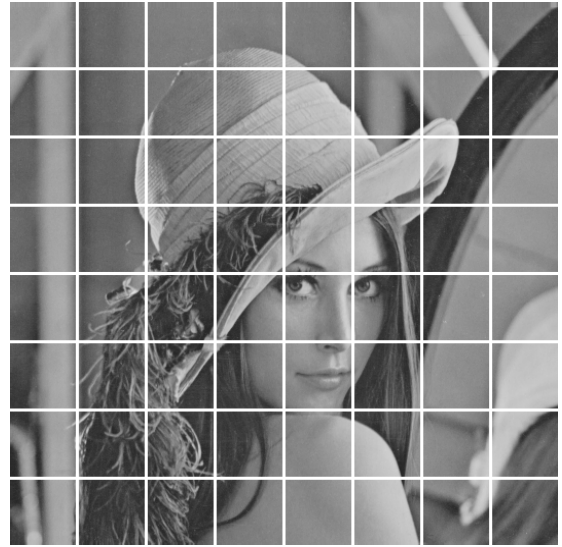

Fig. 7: Original set of sub-blocks of 64-by-64 pixels

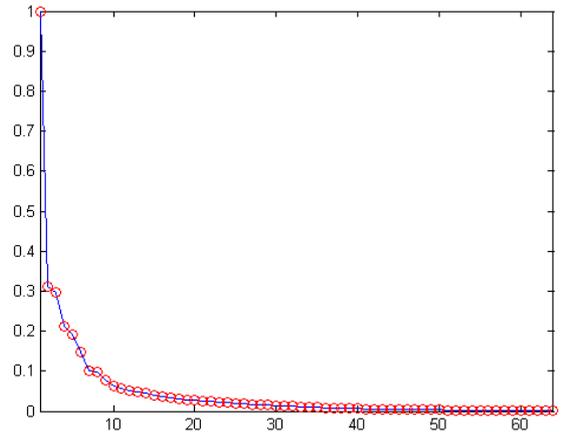

Fig. 8: Eigenvalues spectrum of Fig. 6

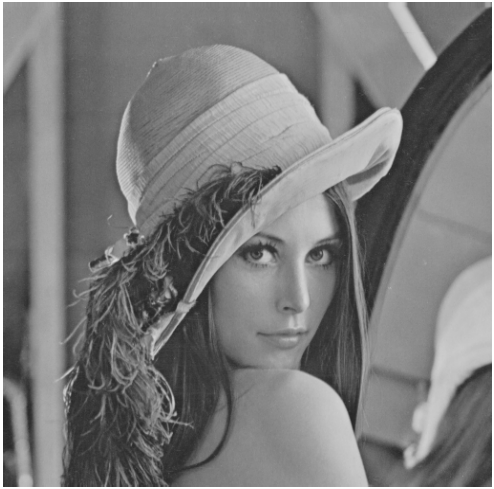

Fig. 6: Lena of 512-by-512 pixels, with 8 bits-per-pixel (bpp)

The Fig.9-10 represents a set of sub-blocks much more efficient than Fig.7-8, because, the sub-blocks of the Fig.7-8 are more correlated morphologically. In Fig.7 is evident than each sub-block represent a little version of Lena. In Fig.10 the first 2 sub-blocks account for about 95% of the total variance, while in Fig.8 the first 46 sub-blocks account for about 95% of the total variance. Therefore, Fig.7 is a inefficient set, while Fig.9 is highly efficient. This is the reason that makes the KLT as efficient in multi and hyperspectral imagery and very inefficient in images alone (monoframe) [21,27-34,38-40,43].

Fig.8 and Fig.10 represents the respective normalized eigenvalues spectrum, i.e., divided by the first eigenvalue (largest).

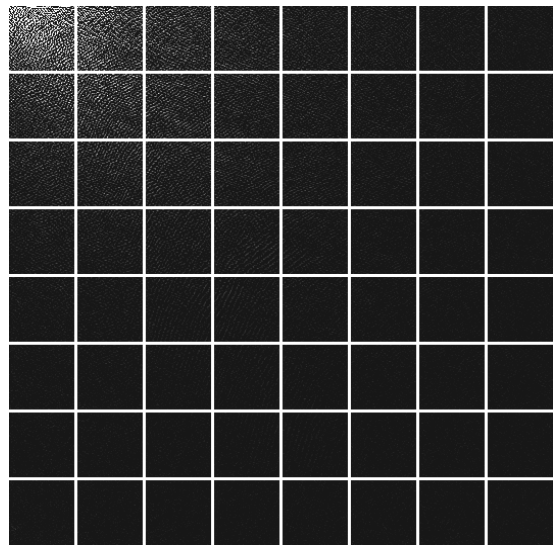

Fig. 9: Efficient set of sub-blocks of 64-by-64 pixels

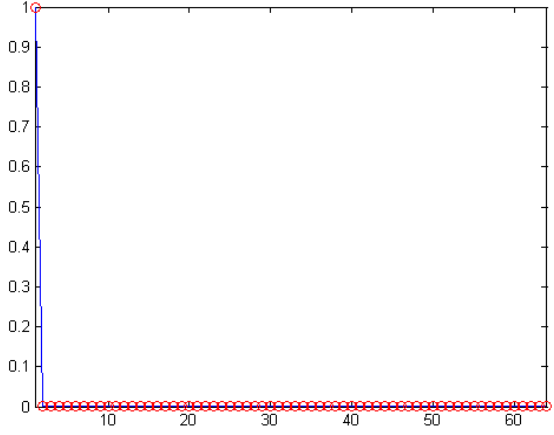

Fig. 10: Eigenvalues spectrum of Fig. 8.

A method prior to KLT (for monoframe images) which resulted in a high correlation of sub-blocks to make the KLT more efficient and will be very welcome.

On the other hands, the inverse KLT will be,

$$\mathbf{x} = \mathbf{V}\,\mathbf{y} + \mathbf{m_x} \tag{9}$$

A complete lossy image compression algorithm based on KLT may be:

**CODEC:**
1. Image sub-blocking with zig-zag scan and construction of three dimensional matrix.
2. KLT to resulting sub-blocks
3. Pruning of sub-blocks based on percentage of resulting covariance matrix
4. Quantization
5. Entropy encoding

*To channel or storage*

**DECODEC:**
6. Entropy decoding
7. Zero-padding: Complete with zeros the sub-blocks pruned
8. Inverse KLT
9. Reconstruction of bidimensional matrix from the new sub-blocks set with inverse of zig-zag scan and image reassembling.

## V. COMBINATIONS

Based on the last section, the proposed solution to achieve the goal is as follows:

*CODEC*
1. **FCT-2D to image**
2. **Image sub-blocking with zig-zag scan and construction of three dimensional matrix**
3. **Construction of three dimensional matrix**
4. **KLT**
5. **Pruning**
6. **Quantization**
7. **Entropy encoding**

*To channel or storage*

*DECODEC*
6. **Entropy decoding**
7. **Zero-padding: Complete with zeros the sub-blocks pruned**
8. **Inverse KLT**
9. **Reconstruction of bidimensional matrix from the new sub-blocks set with inverse of zig-zag scan and image reassembling**
10. **Inverse of FCT-2D**

## VI. METRICS

### A. Data Compression Ratio (CR)

Data compression ratio, also known as compression power, is a computer-science term used to quantify the reduction in data-representation size produced by a data compression algorithm. The data compression ratio is analogous to the physical compression ratio used to measure physical compression of substances, and is defined in the same way, as the ratio between the *uncompressed size* and the *compressed size* [20]:

$$CR = \frac{Uncompressed\ Size}{Compressed\ Size} \tag{10}$$

Thus a representation that compresses a 10MB file to 2MB has a compression ratio of 10/2 = 5, often notated as an explicit ratio, 5:1 (read "five to one"), or as an implicit ratio, 5X. Note that this formulation applies equally for compression, where the uncompressed size is that of the original; and for decompression, where the uncompressed size is that of the reproduction.

### B. Bit-per-pixel (bpp)

The "bits per pixel" refers to the sum of the bits in all three color channels and represents the sum colors available at each pixel before compression ($bpp_{bc}$). However, as a compression metric, the bits-per-pixel refers to the average of the bits in all three color channels, after of compression process ($bpp_{ac}$).

$$bpp_{ac} = \frac{Compressed\ Size}{Uncompressed\ Size} \times bpp_{bc} = \frac{bpp_{bc}}{CR} \tag{11}$$

Besides, bpp is also defined as

$$bpp_{ac} = \frac{Number\ of\ coded\ bits}{Number\ of\ pixels} \qquad (12)$$

*C. Mean Absolute Error (MAE)*

The mean absolute error is a quantity used to measure how close forecasts or predictions are to the eventual outcomes. The mean absolute error (MAE) is given by

$$MAE = \frac{1}{NRxNC} \sum_{nr=0}^{NR-1} \sum_{nc=0}^{NC-1} \| I(nr,nc) - I_d(nr,nc) \| \qquad (13)$$

which for two $NR \times NC$ (rows-by-columns) monochrome images $I$ and $I_d$, where the second one of the images is considered a decompressed approximation of the other of the first one.

*D. Mean Squared Error (MSE)*

The mean square error or MSE in Image Compression is one of many ways to quantify the difference between an original image and the true value of the quantity being decompressed image, which for two $NR \times NC$ (rows-by-columns) monochrome images $I$ and $I_d$, where the second one of the images is considered a decompressed approxi-mation of the other is defined as:

$$MSE = \frac{1}{NRxNC} \sum_{nr=0}^{NR-1} \sum_{nc=0}^{NC-1} \| I(nr,nc) - I_d(nr,nc) \|^2 \qquad (14)$$

*E. Peak Signal-To-Noise Ratio (PSNR)*

The phrase peak signal-to-noise ratio, often abbreviated PSNR, is an engineering term for the ratio between the maximum possible power of a signal and the power of corrupting noise that affects the fidelity of its representation. Because many signals have a very wide dynamic range, PSNR is usually expressed in terms of the logarithmic decibel scale. The PSNR is most commonly used as a measure of quality of reconstruction in image compression, etc [20]. It is most easily defined via the mean squared error (MSE), so, the PSNR is defined as [20]:

$$PSNR = 10 \log_{10}(\frac{MAX_I^2}{MSE}) = 20 \log_{10}(\frac{MAX_I}{\sqrt{MSE}}) \qquad (15)$$

Here, $MAX_I$ is the maximum pixel value of the image. When the pixels are represented using 8 bits per sample, this is 256. More generally, when samples are represented using linear pulse code modulation (PCM) with B bits per sample, maximum possible value of $MAX_I$ is $2^B-1$.

For color images with three red-green-blue (RGB) values per pixel, the definition of PSNR is the same except the MSE is the sum over all squared value differences divided by image size and by three [20].

Typical values for the PSNR in lossy image and video compression are between 30 and 50 dB, where higher is better.

*F. First Gap Percent (FGP) [45]*

This metric is defined as:

$$FGP = \left(1 - \frac{\lambda_2}{\lambda_1}\right) \times 100\% \qquad (16)$$

where $\lambda_2$ is the second eigenvalue, and $\lambda_1$ is the first eigenvalue. Since the spectrum of eigenvalues is monotonically decreasing, if the difference between $\lambda_1$ and $\lambda_2$ is large, this percentual difference should be high, as is the case of Fig.8, and corresponds to Fig.10, where the first normalized eigenvalue is 1, while the second coincides with the axis of abscissae. While in the case of Fig.7, to be different mosaics, this gap is percentual low, so $\lambda_2$ is near $\lambda_1$ as shown in Fig.9. The same figure shows that a large number of eigenvalues have values significantly above zero, which we do not allow efficient compression by pruning, if we get rid of the mosaics that contribute less to the final image at the time of reconstruction [45].

*G. First vs Rest Percent (FRP) [45]*

This metric is defined as:

$$FRP = \left(1 - \frac{\lambda_2 - \lambda_N}{\lambda_1}\right) \times 100\% \qquad (17)$$

and gives us the notion of $\lambda_1$ about the difference between $\lambda_2$ and $\lambda_N$. This metric will be critical to assessing the compression ratio in terms of percentage of pruning the least significant eigenvalues.

*H. First Percent (FP) [45]*

This last metric is defined as:

$$FP = \frac{\lambda_1}{\sum_{i=1}^{N} \lambda_i} \times 100\% \qquad (18)$$

and gives us the notion of the weight of the $\lambda_1$ for the entire spectrum. It will be particularly useful in assessing extreme compression rates [45].

However, the underlying question is: can achieve and artificial state where the last three metrics have values close to 100%?

## VII. COMPUTERS SIMULATIONS

The simulations are organized in two sets of experiments:

*Experiment 1:* KLT vs FCT+KLT
This experiment includes calculations of following metrics:
1. Based on image reconstruction
   1.1. MAE
   1.2. MSE
   1.3. PSNR
2. Based on compression performance
   2.1. CR
   2.2. bpp
   2.3. Elapsed time (etime)
3. Based on spectrum of eigenvalues
   3.1. FGP
   3.2. FRP
   3.3. FP

Main characteristics:
1. Image = Lena
2. Color = gray
3. Size = 512-by-512 pixels
4. Bits-per-pixel = 8
5. Maximum compression rate = 4:1
6. Sub-blocks size = 64-by-64 pixels

*Experiment 2:* FCT+KLT vs JPEG vs JPEG2000
This experiment includes calculations of following metrics:
1. Based on image reconstruction
   1.1. MAE
   1.2. MSE
   1.3. PSNR
2. Based on compression performance
   2.1. CR
   2.2. bpp

Main characteristics:
1. Image = Lena
2. Color = gray
3. Size = 512-by-512 pixels
4. Bits-per-pixel = 8
5. Maximum compression rate = 10:1
6. Sub-blocks size = 32-by-32 pixels for FCT+KLT
               = 8-by-8 pixels for JPEG and JPEG2000

*Experiment 1:* KLT vs FCT+KLT
Based on Fig. 6, which represents to Lena of 512-by-512 pixels, with 8 bits-per-pixel (bpp), Table I shows metrics vs KLT (alone) and combinations of FCT plus KLT.

With identical CR (3.9990) and bpp (2.0005) the rest of metrics shows a great superiority of FCT+KLT in front of KLT alone. In facts, all metrics based on spectrum of eigenvalues demonstrated a marked improvement thanks to the presence of FCT before KLT. Specifically, Fig.11 represents the reconstructed image using KLT alone, while Fig.12 represents the reconstructed image employing FCT before KLT. Look at the block artifacts by not wearing FCT before KLT.

TABLE I: METRICS VS KLT AND FCT+KLT

| Metric | KLT | FCT+KLT |
|---|---|---|
| MAE | 4.1733 | 1.1105 |
| MSE | 39.5500 | 5.3119 |
| PSNR | 32.1593 | 40.8783 |
| CR | 3.9990 | 3.9990 |
| bpp | 2.0005 | 2.0005 |
| etime (seg) | 94.0702 | 93.7234 |
| FGP | 68.7584 | 99.7167 |
| FRP | 68.7854 | 99.7170 |
| FP | 30.8915 | 99.2517 |

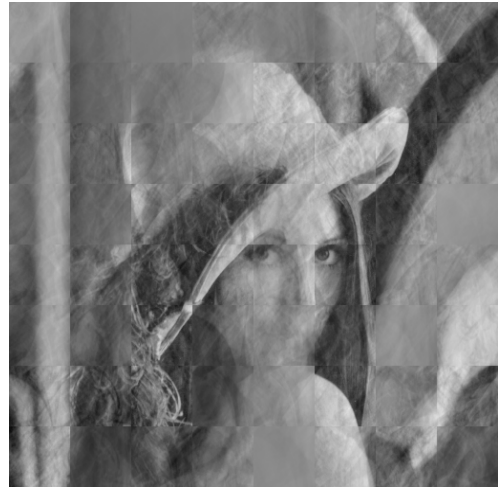

Fig. 11: Reconstructed image using KLT alones.

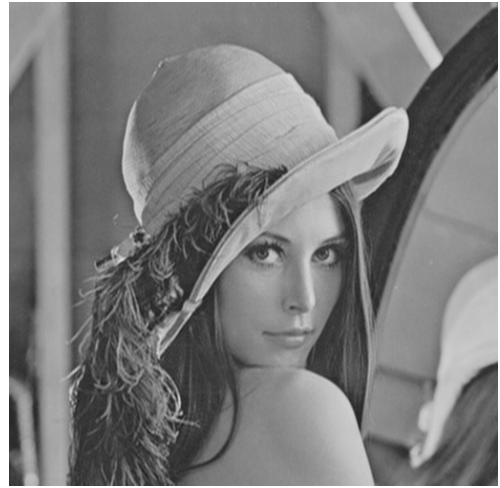

Fig. 12: Reconstructed image using FCT+KLT.

Fig.13 represents the error pixel-to-pixel for KLT alone. Look at the presence of red and blue pixels where the zero value is represented by green. Instead, Fig.14 shows similar values in all pixels, that is to say, green color (representing zero values). This is a clear comparative advantage the novel (FCT+KLT) over the version with KLT alone. This results were already known but with wavelets [43].

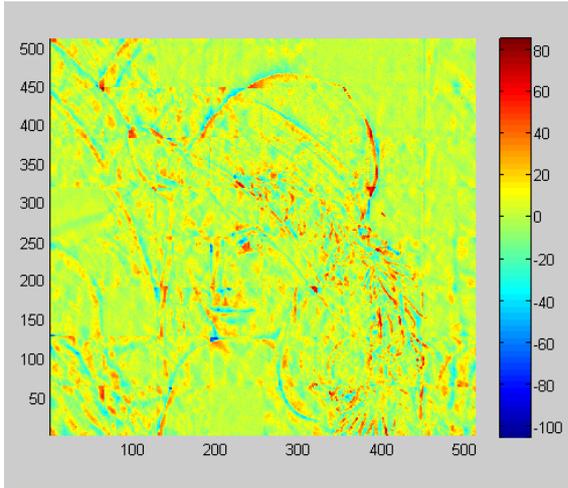

Fig. 13: Error pixel-to-pixel for KLT alone.

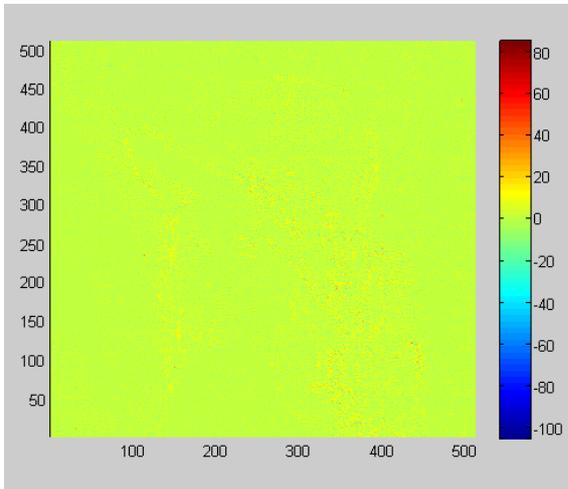

Fig. 14: Error pixel-to-pixel for FCT+KLT.

*Experiment 2:* FCT+KLT vs JPEG vs JPEG2000

Based on Fig.6 too, Table II shows metrics vs JPEG, JPEG2000, and FCT plus KLT.

Though some metrics are better, we must remember that the JPEG and JPEG2000 wears blocks of 8x8 pixels, while, FCT plus KLT wears (in this case) blocks of 32x32. With smaller blocks, we get much higher metric JPEG and JPEG2000.

TABLE II: Metrics vs FCT+KLT, JPEG and JPEG2000

| Metric | FCT+KLT | JPEG | JPEG2000 |
|---|---|---|---|
| MAE | 0.7970 | 0.7510 | 0.8856 |
| MSE | 2.3781 | 2.0604 | 2.8011 |
| PSNR | 44.9242 | 44.9913 | 43.6576 |
| CR | 10.6641 | 4.5453 | 10.0061 |
| bpp | 0.7502 | 1.7600 | 0.7995 |

On the other hand, Fig.15 represents decompressed image using JPEG, Fig.16 represents decompressed image using JPEG2000, and Fig.17 represents reconstructed image using FCT+KLT.

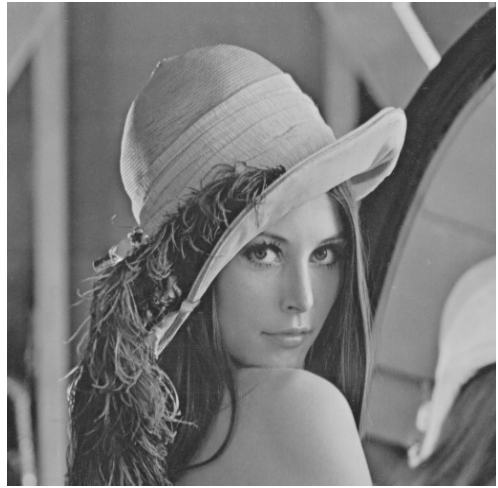

Fig. 15: Decompressed image using JPEG.

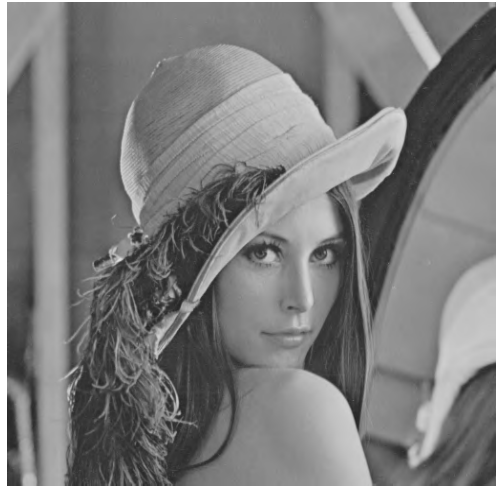

Fig. 16: Decompressed image using JPEG2000.

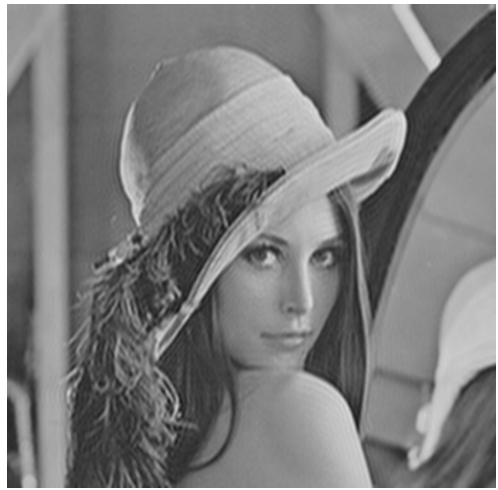

Fig. 17: Reconstructed image using FCT+KLT.

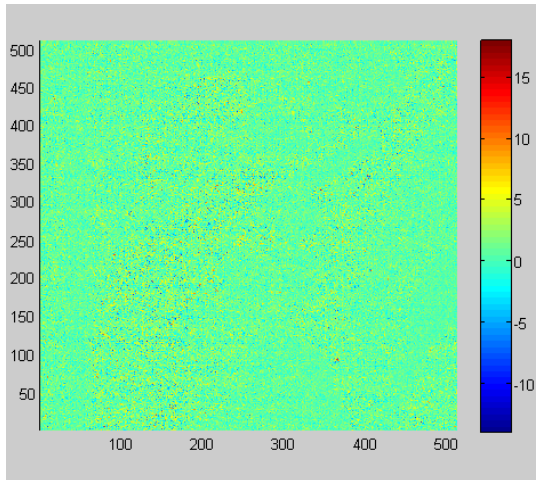

Fig. 18: Error pixel-to-pixel for JPEG.

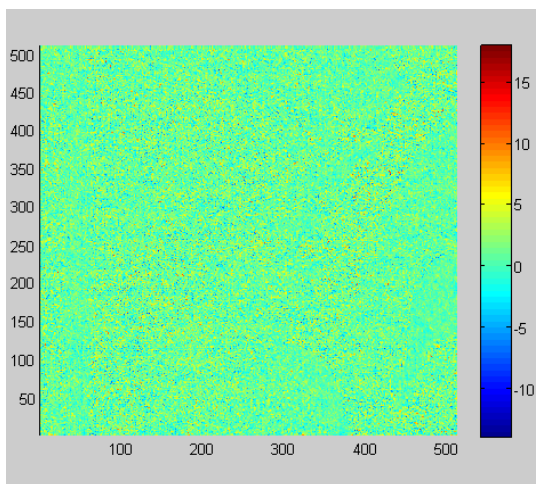

Fig. 19: Error pixel-to-pixel for JPEG2000.

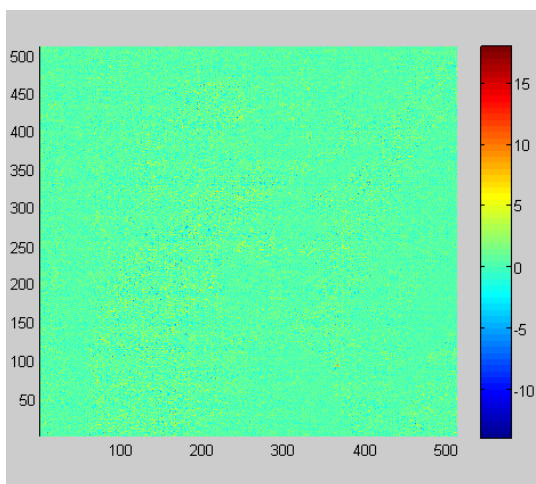

Fig. 20: Error pixel-to-pixel for FCT+KLT.

Figures 18, 19 and 20 represents error pixel-to-pixel for JPEG, JPEG2000, and FCT+KLT respectively.

Finally, all techniques were implemented in MATLAB® (Mathworks, Natick, MA) [46] on a PC with an Intel® Core(TM) QUAD CPU Q6600 2.40 GHz processors and 4 GB RAM.

## VIII. CONCLUSION

*Experiment 1:* KLT vs FCT+KLT

In this experiment FCT+KLT is better than KLT alone.

As shown in the Figure 11, although KLT is optimum, it is inefficient in the sub-blocks decorrelation, in the cases where such sub-blocks are morphologically differents. The experimental evidence shows that previous FCT supplies KLT of the necessary morphological affinity, see Figure 12.

As discussed earlier, the KLT is theoretically the optimum method to spectrally decorrelate a set of sub-blocks image. However, it is computationally expensive. Future research should be geared to the use of lower-cost computational approaches [43,45].

*Experiment 2:* FCT+KLT vs JPEG vs JPEG2000

In this experiment, we have demonstrated than FCT+KLT have the same CR than JPEG2000 but with blocks of 32-by-32 pixels vs JPEG2000 with blocks of 8-by-8 pixels. These, represents a faster encoding/decoding process and a smaller number of blocks to be manipulated.

As seen in the Table II, FCT+KLT metrics are among the values of the metrics of JPEG and JPEG2000, with similar error pixel-to-pixel (see Figures 18, 19 and 20).

Finally, the reconstructed images have a similar look-and-feel in the three cases (see Figures 15, 16 and 17).